\documentclass[10pt,twocolumn,letterpaper]{article}

\usepackage[pagenumbers]{cvpr} \newif\ifSatThreeDAppendix
\SatThreeDAppendixtrue
\def\SatThreeDMainDocument{1}
\usepackage{xr-hyper}
\usepackage{times,graphicx,amsmath,amssymb,booktabs,tabularx,array}
\usepackage{xcolor}
\usepackage{cuted,caption}
\usepackage{stfloats}
\usepackage{adjustbox}
\usepackage{makecell}
\usepackage{pifont}
\usepackage{multirow}
\newcommand{\cmark}{\textcolor{green!80!black}{\ding{51}}}
\newcommand{\xmark}{\textcolor{red!95!black}{\ding{55}}}
\newcommand{\pmark}{\textcolor{orange!85!black}{$\triangle$}}
\usepackage[normalem]{ulem}

\newcommand{\red}[1]{{\color{red}#1}}
\newcommand{\todo}[1]{{\color{red}#1}}
\newcommand{\TODO}[1]{\textbf{\color{red}[TODO: #1]}}

\definecolor{cvprblue}{rgb}{0.21,0.49,0.74}
\usepackage[pagebackref,breaklinks,colorlinks,allcolors=cvprblue]{hyperref}

\ifSatThreeDAppendix
\else
\fi

\def\confName{CVPR}
\def\confYear{2027}

\title{Sat3D-Bench: Benchmarking Large-Scale 3D Scene Generation \\ from a Single Satellite Image}

\author{
\textbf{Congyi Fan$^1$},
\textbf{Dezhi Zhu$^{1}$}, 
\textbf{Youtian Lin$^2$},
\textbf{Xuanjia Zhao$^3$},\\
\textbf{Pengming Feng$^4$},
\textbf{Guangjun He$^4$},
\textbf{Haiwei Pan$^1$},
\textbf{Jian Guan$^{1}$\thanks{Corresponding author: j.guan@hrbeu.edu.cn}}
\\ \\
$^1$Harbin Engineering University \quad $^2$Nanjing University \quad $^3$University of Surrey\\
$^4$State Key Laboratory of Space Information System and Integrated Application \\
}

\usepackage{iftex}
\ifPDFTeX
  \usepackage[T1]{fontenc}
\else
  \usepackage{fontspec}
\fi
\usepackage{microtype}

\begin{document}

\maketitle

\begin{abstract}
Generating a large-scale 3D urban scene from a single satellite image requires preserving scene-wide layout, surface appearance, and relative vertical structure. Visual plausibility alone does not establish fidelity to the corresponding environment, motivating evaluation against paired 3D references. 
We introduce Sat3D-Bench, a benchmark comprising 550 high-quality pairs of satellite images and textured 3D urban scenes curated from approximately 56K candidate 3D scenes. The selected scenes span 29 cities across five continents, with scene extents ranging from 100 to 1,000 m. Predictions and references are independently normalized and evaluated using fixed-view renderings, complete mesh surfaces, and relative-elevation maps. Eleven metrics are organized into appearance quality and geometry fidelity, covering visual quality, reference appearance agreement, surface distance, matching and coverage, input--shape alignment, and relative elevation. 
We systematically evaluate 22 generation methods to characterize their ability to recover large-scale urban scenes in terms of appearance, surface completeness, and relative vertical structure. Sat3D-Bench establishes a common foundation for method comparison and capability diagnosis, supporting research on single-image 3D generation for complex urban environments. Project page: \href{https://geox-research.github.io/Sat3D-Bench/}{Sat3D-Bench}.
\end{abstract}

\begin{figure}[t]
    \centering
    \vspace{-6pt}
    \includegraphics[width=.93\linewidth]{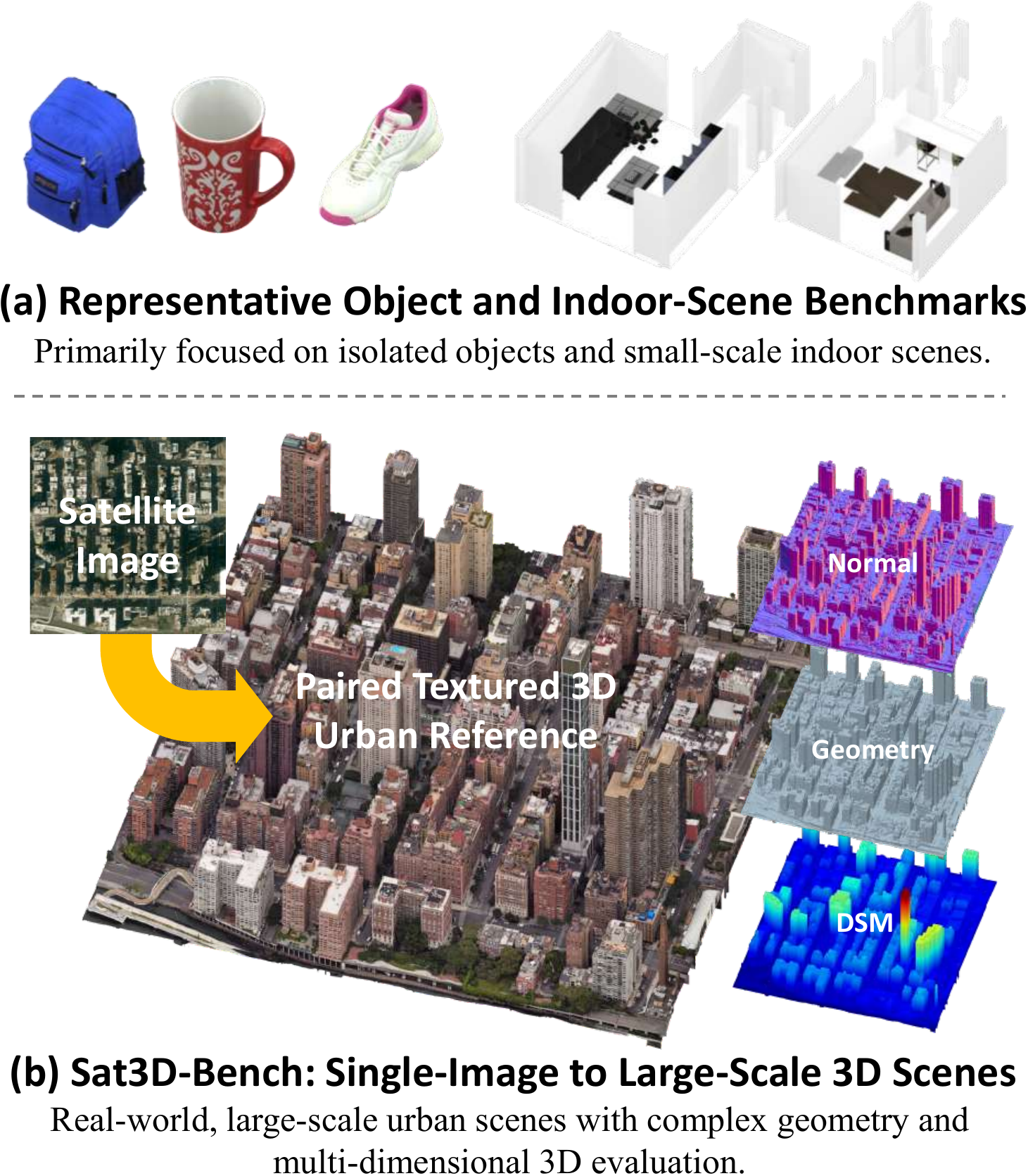}
   \caption{\textbf{Evaluating large-scale urban scene generation.}
(a) Representative evaluation resources cover individual objects~\cite{downs2022gso} and indoor scenes~\cite{fu2021front}.
(b) Sat3D-Bench pairs satellite images with textured 3D urban references, visualized through texture, surface normals, geometry, and relative elevation. These paired references support complementary evaluation of generated appearance and 3D structure.}
    \label{fig:teaser}
    \vspace{-12pt}
\end{figure}

\section{Introduction}
\label{sec:intro}

Generating a large-scale 3D scene from a single satellite image offers a way to model urban environments from overhead observations. More broadly, 3D environments support embodied interaction~\cite{deitke2022procthor}, simulation~\cite{dosovitskiy2017carla}, and urban digital-twin modeling~\cite{wysocki2026tum2twin}. Unlike isolated-object generation, this task requires preserving the spatial arrangement and relative proportions of buildings, roads, and surrounding structures while inferring facade textures and 3D geometry from limited visual evidence. A generated scene may appear visually plausible while differing from the corresponding environment in building layout, surface appearance, or relative height. Evaluating these results therefore requires assessing both visual quality and fidelity to the corresponding environment.

Existing 3D datasets cover both individual objects and multi-structure scenes. Object-level datasets, such as Pix3D~\cite{sun2018pix3d}, Google Scanned Objects~\cite{downs2022gso}, and OmniObject3D~\cite{wu2023omniobject3d}, provide references for evaluating object shape and appearance. Scene-level datasets capture multiple structures and their spatial relationships.  These include indoor environments in 3D-FRONT~\cite{fu2021front} and ScanNet++~\cite{yeshwanth2023scannetpp}, as well as outdoor urban environments in UrbanScene3D~\cite{urbanscene3d}, MatrixCity~\cite{li2023matrixcity}, and H3D~\cite{kolle2021h3d}. The latter primarily support reconstruction, neural rendering, and semantic segmentation, respectively. 
Satellite-street-view pairs and DSMs restrict reference-based evaluation to observed views or height fields, leaving full-surface geometry and texture insufficiently assessed~\cite{qian2023sat2density,sat2densitypp,sat3dgen}. These data settings motivate a common evaluation framework that jointly assesses scene appearance, surface geometry, and relative elevation, as illustrated in Figure~\ref{fig:teaser} and summarized in Table~\ref{tab:benchmark_comparison}.

Relevant 3D generation methods span general image-to-3D generation, scene and world generation, and satellite-to-3D generation. General image-to-3D methods, such as TripoSR~\cite{tochilkin2024triposr} and TRELLIS~\cite{xiang2025trellis}, generate 3D assets from a single image. Scene and world generation methods, including SceneGen~\cite{meng2025scenegen} and WonderWorld~\cite{yu2025wonderworld}, extend generation to multi-object scenes and larger environments. Satellite-to-3D methods, such as Sat2Density~\cite{qian2023sat2density} and Sat3DGen~\cite{sat3dgen}, target urban environments conditioned on satellite imagery. Their differing generation targets and conditioning inputs motivate a systematic comparison of their capabilities in large-scale 3D scene generation from a single satellite image under explicit input conditions and common evaluation criteria. However, a comprehensive benchmark is still needed to compare these categories across scene scales and complexity levels.

To this end, we introduce \textbf{Sat3D-Bench} for evaluating large-scale 3D scene generation from a single satellite image. We curate \textbf{550 high-quality pairs of satellite images and textured 3D urban scenes} from approximately
\textbf{56K candidate 3D scenes} through automated checks and manual review. 
The selected pairs span \textbf{29 cities across five continents}, with scene extents ranging from \textbf{100 to 1,000 m}. 
These pairs provide references for assessing how faithfully generated scenes reproduce the corresponding environments. Building on these paired data, we establish a common evaluation framework with shared scene normalization and observation settings. Eleven metrics are organized into two groups, i.e., appearance quality and geometry fidelity. Appearance evaluation measures visual quality and reference appearance agreement, while geometry evaluation measures surface distance, matching and coverage, input--shape alignment, and relative elevation. We further analyze generation performance across scene scales and complexity levels. 

Using Sat3D-Bench, we systematically evaluate \textbf{22 generation methods} spanning satellite-to-3D, general image-to-3D, and scene and world generation. The evaluation characterizes their ability to recover large-scale urban scenes in terms of appearance, surface completeness, and relative vertical structure. These comparisons establish a common foundation for capability diagnosis and identify priorities for advancing single-image 3D generation in complex urban environments.

Our main contributions are as follows:
\begin{itemize}
    \item We introduce Sat3D-Bench, comprising 550 high-quality  pairs of satellite images and textured 3D urban scenes from 29 cities across five continents, providing references for evaluating large-scale 3D scene generation from a single satellite image.

    \item We establish an evaluation framework with 11 metrics covering appearance quality and geometry fidelity under shared scene normalization and observation settings, supporting analysis across scene scales and complexity levels.

    \item We systematically evaluate 22 generation methods and show that higher visual quality does not necessarily imply greater reference fidelity, while similar mean surface distances can accompany differences in surface coverage and geometric agreement.
\end{itemize}

\begin{table*}[t]
\centering
\providecommand{\cmark}{\textcolor{green!80!black}{\ding{51}}}
\providecommand{\xmark}{\textcolor{red!95!black}{\ding{55}}}
\providecommand{\pmark}{\textcolor{orange!90!black}{\ensuremath{\triangle}}}

\caption{\textbf{Comparison of representative 3D datasets and evaluation settings.} Object-level and scene-level resources are compared in terms of data characteristics, evaluation settings, and public data availability. \cmark\ indicates documented support, \xmark\ indicates support not established in the cited work or official release, and \pmark\ denotes partial or mixed support. Public data may require registration or research-use approval.}
\label{tab:benchmark_comparison}

\begingroup
\setlength{\tabcolsep}{3.0pt}
\renewcommand{\arraystretch}{1.0}
\small

\begin{adjustbox}{max width=\textwidth}
\begin{tabular}{@{}l l c *{6}{c}@{}}
\toprule

\multirow[c]{3}{*}{\textbf{Dataset / Benchmark}} &
\multirow[c]{3}{*}{\textbf{Size}} &
\multirow[c]{3}{*}{\makecell{\textbf{Public Data}}} &
\multicolumn{3}{c}{\textbf{Data Characteristics}} &
\multicolumn{3}{c}{\textbf{Evaluation Settings}} \\

\cmidrule(lr){4-6}
\cmidrule(lr){7-9}

& & &
\multirow[c]{2}{*}{\textbf{Textured 3D}} &
\textbf{Large Outdoor} &
\textbf{Satellite--3D} &
\textbf{Appearance} &
\textbf{Geometry} &
\textbf{Scene Complexity} \\

& & & &
\textbf{Scenes} &
\textbf{Pairs} &
\textbf{Eval.} &
\textbf{Eval.} &
\textbf{Levels} \\

\midrule
\multicolumn{9}{@{}l}{\emph{\textbf{A. Object-Level Datasets}}} \\

ShapeNetCore~\cite{chang2015shapenet}
& 51.3K objects
& \cmark
& \cmark & \xmark & \xmark
& \xmark & \xmark & \xmark \\

Pix3D~\cite{sun2018pix3d}
& 395 objects
& \cmark
& \xmark & \xmark & \xmark
& \xmark & \cmark & \xmark \\

Google Scanned Objects~\cite{downs2022gso}
& 1,030 objects
& \cmark
& \cmark & \xmark & \xmark
& \xmark & \xmark & \xmark \\

CO3D~\cite{reizenstein2021co3d}
& 18,619 videos
& \cmark
& \xmark & \xmark & \xmark
& \cmark & \cmark & \xmark \\

OmniObject3D~\cite{wu2023omniobject3d}
& 6,000 objects
& \cmark
& \cmark & \xmark & \xmark
& \cmark & \cmark & \xmark \\

ABO~\cite{collins2022abo}
& 7,953 objects
& \cmark
& \cmark & \xmark & \xmark
& \cmark & \cmark & \xmark \\

3D-FUTURE~\cite{fu20213d}
& 9,992 objects
& \cmark
& \cmark & \xmark & \xmark
& \cmark & \cmark & \xmark \\

Objaverse~\cite{deitke2023objaverse}
& 800K+ assets
& \cmark
& \cmark & \pmark & \xmark
& \xmark & \xmark & \xmark \\

Objaverse-XL~\cite{deitke2023objaversexl}
& 10M+ assets
& \cmark
& \pmark & \pmark & \xmark
& \cmark & \xmark & \xmark \\

\midrule
\multicolumn{9}{@{}l}{\emph{\textbf{B. Scene-Level Datasets}}} \\

3D-FRONT~\cite{fu2021front}
& 18,968 rooms
& \cmark
& \cmark & \xmark & \xmark
& \cmark & \cmark & \xmark \\

ScanNet++~\cite{yeshwanth2023scannetpp}
& 460 scenes
& \cmark
& \xmark & \xmark & \xmark
& \cmark & \xmark & \xmark \\

UrbanScene3D~\cite{urbanscene3d}
& 16 scenes
& \cmark
& \cmark & \cmark & \xmark
& \xmark & \cmark & \xmark \\

MatrixCity~\cite{li2023matrixcity}
& 2 scenes
& \cmark
& \cmark & \cmark & \xmark
& \cmark & \xmark & \xmark \\

H3D~\cite{kolle2021h3d}
& 1 scene
& \cmark
& \cmark & \cmark & \xmark
& \xmark & \xmark & \xmark \\

DFC2019~\cite{dfc2019}
& 131 tiles
& \cmark
& \xmark & \cmark & \cmark
& \xmark & \cmark & \xmark \\

Sat2City v2~\cite{sat2cityv2}
& 1,590 pairs
& \xmark
& \cmark & \cmark & \cmark
& \cmark & \cmark & \xmark \\

\textbf{Sat3D-Bench (Ours)}
& \textbf{550 scenes}
& --
& \cmark & \cmark & \cmark
& \cmark & \cmark & \cmark \\

\bottomrule
\end{tabular}
\end{adjustbox}

\begin{minipage}{\textwidth}
\footnotesize

\end{minipage}
\endgroup
\vspace{-10pt}
\end{table*}

\section{Related Work}
\label{sec:related}

\subsection{Object and Scene Datasets}

\noindent\textbf{Object-Level Datasets.} 
Object-level 3D datasets provide geometry and appearance references for individual assets.   ShapeNetCore~\cite{chang2015shapenet} and 3D-FUTURE~\cite{fu20213d} provide CAD assets, while Google Scanned Objects~\cite{downs2022gso} and OmniObject3D~\cite{wu2023omniobject3d} provide scans of real objects. Pix3D~\cite{sun2018pix3d} and ABO~\cite{collins2022abo} pair object images with 3D models, and CO3D~\cite{reizenstein2021co3d} supplies multi-view images and reconstructed geometry. Objaverse~\cite{deitke2023objaverse} and Objaverse-XL~\cite{deitke2023objaversexl} further expand the scale and diversity of 3D asset collections. These resources support object-level generation and reconstruction, while urban scene generation additionally requires preserving the spatial layout and relative proportions of multiple structures.

\noindent\textbf{Scene-Level Datasets.}  
Scene-level 3D datasets extend reference coverage to multiple objects and their surrounding environments. 3D-FRONT~\cite{fu2021front} and ScanNet++~\cite{yeshwanth2023scannetpp} provide furnished synthetic interiors and scanned indoor environments, respectively. At the outdoor urban scale, UrbanScene3D~\cite{urbanscene3d}, MatrixCity~\cite{li2023matrixcity}, and H3D~\cite{kolle2021h3d} primarily support reconstruction, neural rendering, and semantic segmentation, respectively. 
Satellite--street-view pairs used by Sat2Density~\cite{qian2023sat2density} and Sat2Density++~\cite{sat2densitypp}, together with the DSM references introduced by Sat3DGen~\cite{sat3dgen}, lack complete textured 3D references for full-surface geometry and texture evaluation.    
Sat2City~\cite{sat2city} uses synthetic height-map--3D pairs, while Sat2City v2~\cite{sat2cityv2} introduces real satellite images paired with textured meshes and evaluates geometry and satellite-to-asset feature alignment. 
Sat3D-Bench broadens the comparison to 22 methods across satellite-to-3D, general image-to-3D, and scene and world generation under shared evaluation settings. It jointly assesses reference appearance, surface recovery, and relative elevation across scene complexity and scale. Table~\ref{tab:benchmark_comparison} summarizes the data characteristics and evaluation settings of representative resources.

\subsection{3D Generation Methods}

\noindent\textbf{General Image-to-3D Generation.}  
General image-to-3D methods generate 3D assets from a single image. LRM~\cite{hong2024lrm}, TripoSR~\cite{tochilkin2024triposr}, InstantMesh~\cite{xu2024instantmesh}, and CRM~\cite{wang2024crm} adopt feed-forward reconstruction, with the latter two using generated multi-view images. Direct3D~\cite{wu2024direct3d}, Direct3D-S2~\cite{wu2025direct3ds2}, TRELLIS~\cite{xiang2025trellis}, TRELLIS.2~\cite{xiang2026trellis2}, and TripoSG~\cite{li2025triposg} learn 3D generative priors through latent diffusion or flow models. For image conditioning,  Michelangelo~\cite{zhao2023michelangelo} aligns shape, image, and text representations, Pixal3D~\cite{li2026pixal3d} introduces pixel-aligned image conditioning, and Hi3DGen~\cite{ye2025hi3dgen} guides geometry generation with predicted normals. Hunyuan3D-2.1~\cite{hunyuan3d2025v21} and Step1X-3D~\cite{li2025step1x3d} combine geometry generation with texture synthesis, while SAM 3D Objects~\cite{sam3dteam2025sam3d3dfyimages} predicts object geometry, texture, and layout. For satellite-conditioned urban generation, evaluation should further assess whether multiple structures and their spatial relationships are preserved.

\noindent\textbf{Scene and World Generation.} 
Scene and world generation methods produce multi-object scenes or larger environments conditioned on visual or textual inputs. SceneGen~\cite{meng2025scenegen} generates scenes in a single feed-forward pass, WonderWorld~\cite{yu2025wonderworld} supports interactive scene expansion, and EvoScene~\cite{zheng2025self} iteratively refines scenes through 2D--3D feedback. SynCity~\cite{engstler2025syncity} combines pretrained image and 3D generators, and our comparison uses its single-image reconstruction workflow. Extend3D~\cite{yoon2026extend3d} extends object-centric latent priors to larger spatial domains, bridging asset and scene generation. Related research explores compositional layouts~\cite{paschalidou2021atiss,huang2025midi,tian2026sceneregen,sun2026harmony}, single-image scene expansion~\cite{yenphraphai2026rome,liang2025wonderland}, and city synthesis~\cite{lin2023infinicity,xie2024citydreamer,xie2025gaussiancity}. Scalable 3D priors~\cite{liu2026earthcrafter,ren2024xcube} and sparse-view reconstruction~\cite{ren2024scube} further support scene modeling. As these approaches broaden the spatial extent of generated environments, evaluation should consider local appearance together with overall layout and structural completeness.

\noindent\textbf{Satellite-to-3D Generation.} 
Satellite-to-3D methods generate corresponding 3D urban environments conditioned on satellite imagery. Sat2Density~\cite{qian2023sat2density} learns density from satellite--ground image pairs, Sat2Density++~\cite{sat2densitypp} improves street-view synthesis by modeling street-view-specific appearance, and Sat3DGen~\cite{sat3dgen} targets street-level 3D scene generation. Related approaches explore satellite-conditioned scene synthesis~\cite{sat2scene,sat2city,sat2cityv2}, building and urban modeling~\cite{jin2025satskylines,wang2025sat2realcity,cao2026metaearth3d}, and Gaussian or point-based generation~\cite{qian2026gsvoxel,wang2026gridflow}. Because satellite imagery depicts a specific urban area, evaluation should assess both visual quality and fidelity to that area's appearance and structure. Sat3D-Bench uses paired textured urban references and common evaluation criteria to compare appearance quality and geometry fidelity across the three method families in Table~\ref{tab:Sat3D-Bench_4d_height2_full}.

\begin{figure*}[t]
    \centering
    \includegraphics[width=.93\linewidth]{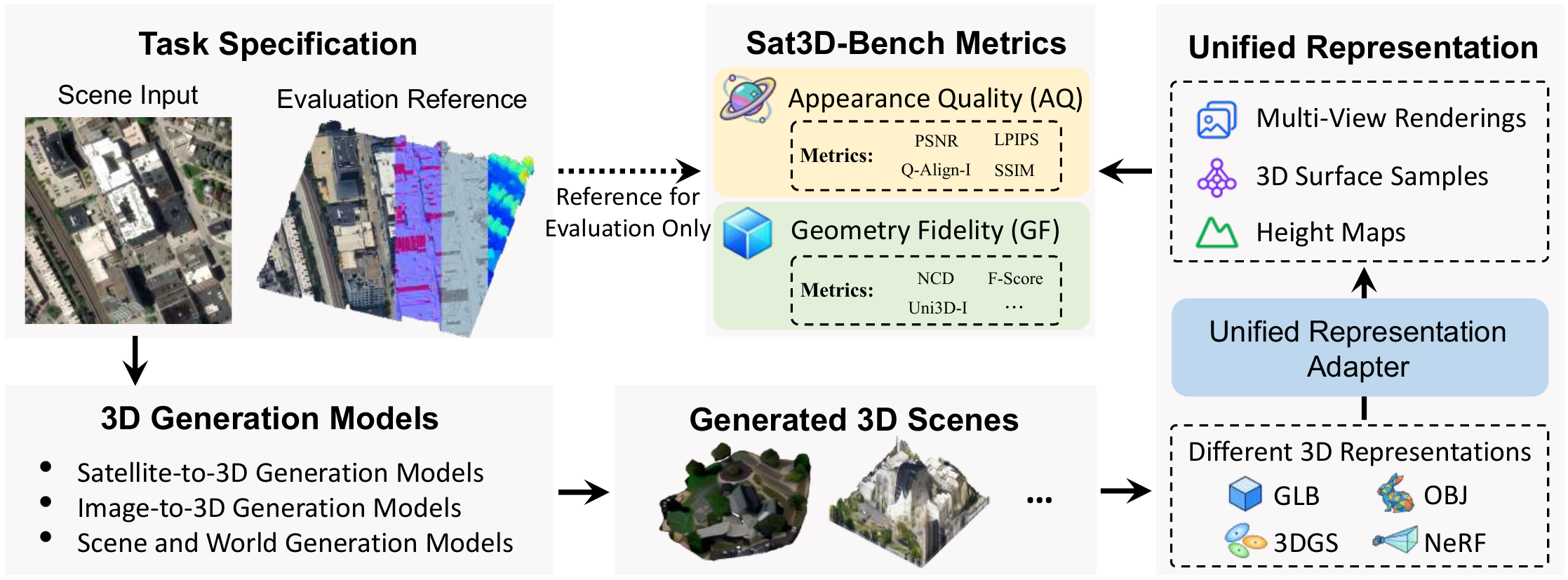}
    \vspace{-4pt}
    \caption{\textbf{Overview of Sat3D-Bench.} Given a single satellite image, 3D generation models produce scenes in different representations. A unified representation adapter converts these outputs into unified representations, including multi-view renderings, 3D surface samples, and height maps. The generated scenes are then evaluated using Sat3D-Bench metrics grouped into Appearance Quality (AQ) and Geometry Fidelity (GF).}
    \label{fig:overview}
    \vspace{-13pt}
\end{figure*}

\section{Sat3D-Bench}
\label{sec:benchmark}

This section presents the overall design of Sat3D-Bench. We first define the generation task and establish common scene-processing rules in Section~\ref{sec:task}. We then describe dataset construction and scene curation in Section~\ref{sec:curation}, followed by appearance and geometry evaluation in Section~\ref{sec:metrics}. Figure~\ref{fig:overview} illustrates the overall pipeline.

\subsection{Task Definition}
\label{sec:task}

\noindent\textbf{Single-Image Scene Generation.}
The task is to generate a textured 3D scene from a single satellite image while preserving the corresponding environment's spatial layout, surface appearance, and relative vertical structure. Let $i$ index scenes and $j$ index generation methods. For scene $i$, the generation model $\mathcal G_j$ takes an RGB satellite image $\mathbf I_i\in\mathbb R^{H\times W\times3}$, where $H$ and $W$ denote image height and width and the three channels correspond to RGB:
\begin{equation}
\widehat{\mathbf S}_{ij}=\mathcal G_j(\mathbf I_i).
\label{eq:task}
\end{equation}
Here, $\widehat{\mathbf S}_{ij}$ denotes the 3D output generated by method $j$ for scene $i$. Its paired textured reference scene, $\mathbf S_i^{\mathrm{ref}}$, is reserved for evaluation.

\noindent\textbf{Unified Representation Adapter.}
To evaluate different 3D representations within a common representation and spatial conventions, we introduce a Unified Representation Adapter, as shown in Figure~\ref{fig:overview}. Its method-specific operation $\mathcal A_j$ converts outputs to meshes where needed and standardizes their orientation using fixed rules. A common normalization operator $\mathcal N$ then independently normalizes the adapted prediction and its reference to support comparison of internal proportions and relative structure:
\begin{equation}
\widetilde{\mathbf S}_{ij}
=\mathcal N\!\left(\mathcal A_j(\widehat{\mathbf S}_{ij})\right),\qquad
\widetilde{\mathbf S}_i^{\mathrm{ref}}
=\mathcal N(\mathbf S_i^{\mathrm{ref}}).
\label{eq:independent_scene_norm}
\end{equation}
Here, $\widetilde{\mathbf S}_{ij}$ and $\widetilde{\mathbf S}_i^{\mathrm{ref}}$ denote the normalized prediction and reference meshes used for evaluation, respectively. The operator $\mathcal N$ centers each mesh horizontally, shifts its minimum height to zero, and applies isotropic scaling so that its maximum horizontal extent equals one. The operation $\mathcal A_j$ is fixed for each method and is not adjusted using individual reference scenes. Mesh conversion, orientation, and normalization details are provided in Section~\ref{sec:supp_adapters} of the supplementary material.

\begin{figure}[t]
    \centering
    \includegraphics[width=.93\linewidth]{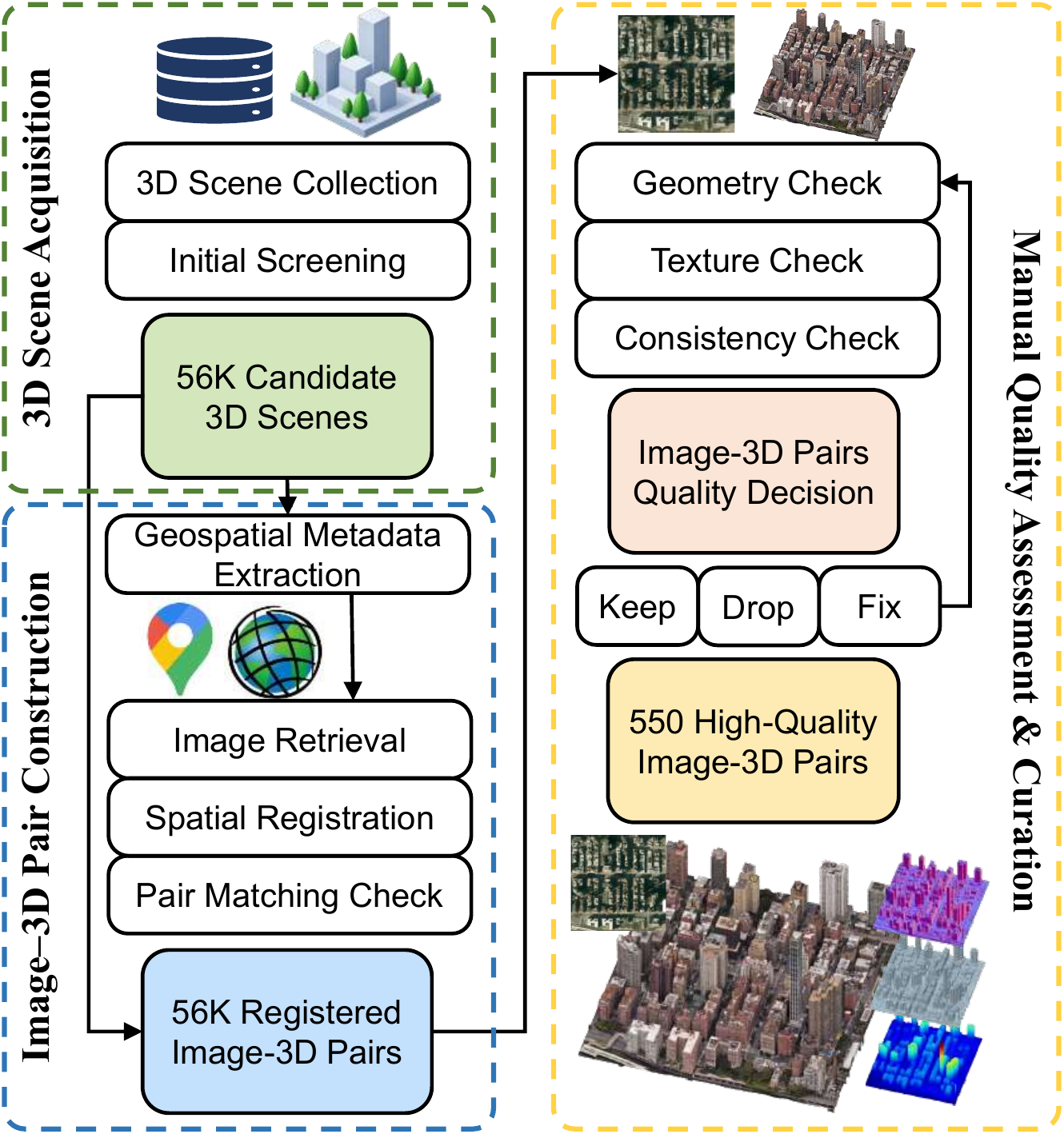}
    \caption{\textbf{Sat3D-Bench collection and curation pipeline.} Starting from 56K registered image--3D pairs, we curate 550 high-quality evaluation pairs through urban-scene selection and quality checks.}
    \label{fig:data_pipeline}
    \vspace{-16pt}
\end{figure}

\begin{figure*}[t]
    \centering
    \includegraphics[width=.95\textwidth]{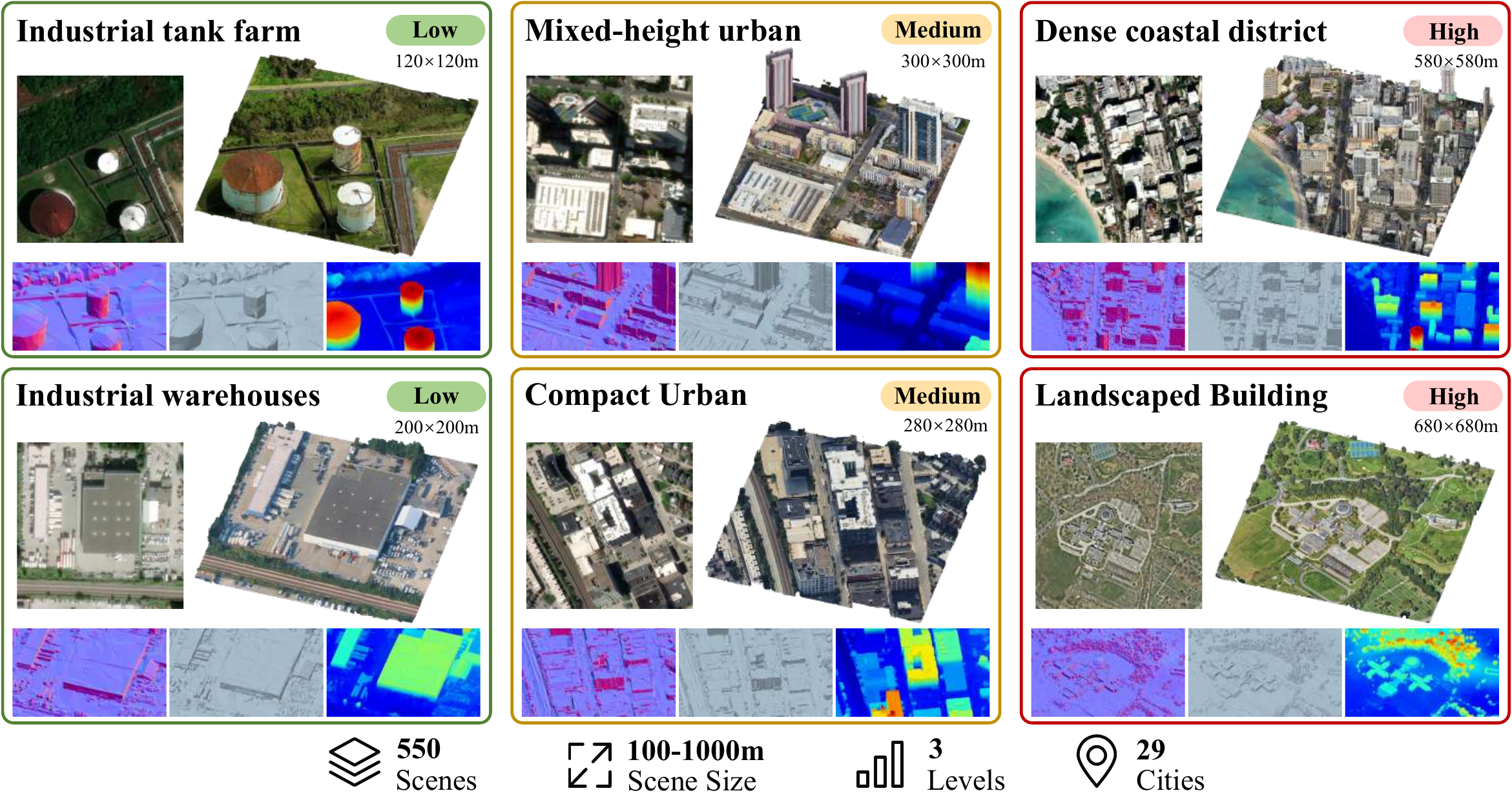}
    \vspace{-6pt}
    \caption{\textbf{Representative Sat3D-Bench scenes.} Six examples from the Low, Medium, and High scene complexity groups, defined by the composite score $\kappa_i$ in Section~\ref{sec:curation}, illustrate varied urban layouts and crop scales. Each shows an overhead image, textured 3D reference, normals, geometry, and surface-elevation visualization. The dataset comprises 550 high-quality scenes from 29 cities.}
    \label{fig:data_demo}
    \vspace{-8pt}
\end{figure*}

\subsection{Dataset Construction and Scene Grouping}
\label{sec:curation}
Sat3D-Bench comprises 550 high-quality evaluation pairs curated from approximately 56K candidate 3D scenes. These pairs span 29 cities across five continents. As shown in Figure~\ref{fig:data_pipeline}, dataset construction consists of scene acquisition, image--3D pairing, and quality assessment. Figure~\ref{fig:data_demo} illustrates their layouts, appearance, and vertical structure.

\noindent\textbf{Scene Acquisition and Pairing.}
Candidate scenes are obtained through photogrammetric reconstruction from oblique aerial imagery~\cite{frommholz2016oblique,michel2020new}, manual modeling with Blender~\cite{blender}, and geospatial data-assisted construction with Blosm~\cite{blosm}. Initial screening yields approximately 56K candidate scenes. Geospatial metadata supports retrieval of the corresponding overhead imagery, followed by spatial registration and pair-matching checks. Crop extent is retained for scale-based analysis.

\noindent\textbf{Reference Quality Control.}
Quality control checks the usability of each reference and its correspondence with the paired image. Automated checks cover file readability, finite coordinates, texture availability, crop extent, and raster validity. Manual review examines scene coverage, truncated structures, texture quality, and image--scene correspondence. Candidates are retained, rejected, or corrected, with corrected samples undergoing the same review to form the final 550 evaluation pairs. Source identifiers, attribution, and applicable licenses are recorded in the release metadata.

\begin{figure}[t]
    \centering
    \includegraphics[width=.95\linewidth]{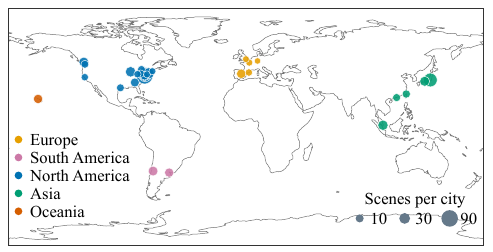}
    \vspace{-8pt}
    \caption{\textbf{Geographic distribution of Sat3D-Bench.} Colors denote geographic regions, and marker sizes indicate the number of scenes per city.}
    \label{fig:geography}
    \vspace{-15pt}
\end{figure}

\noindent\textbf{Geographic and Structural Coverage.}
The collection covers industrial areas, residential neighborhoods, and dense urban blocks with varied building coverage and height relief. As shown in Figure~\ref{fig:geography}, North America, Asia, South America, Europe, and Oceania contribute 329, 128, 40, 34, and 19 scenes, respectively. Overall results weight scenes equally and therefore reflect this scene distribution rather than equal weighting of continents.

\begin{figure}[t]
    \centering
    \includegraphics[width=.93\linewidth]{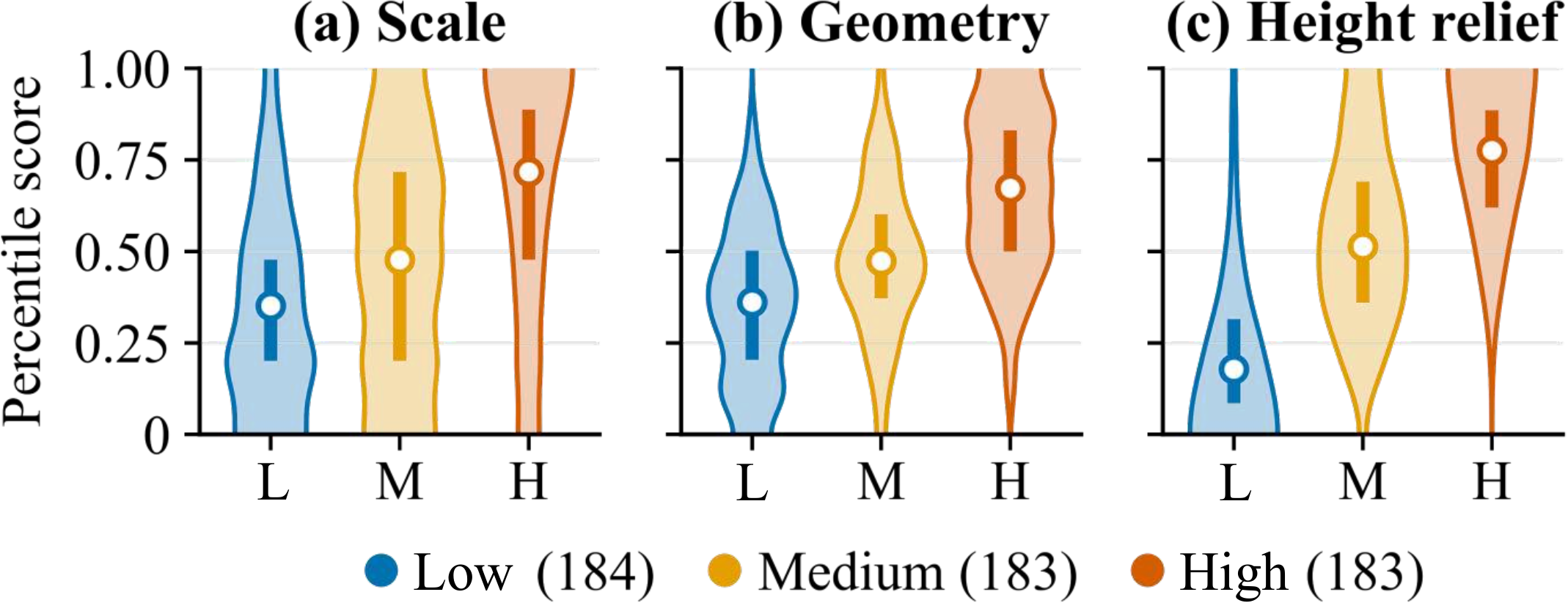}
    \vspace{-8pt}
    \caption{\textbf{Scene complexity groups.} Violin plots show percentile-score distributions of (a) scene scale, (b) geometry, and (c) height relief across Low, Medium, and High scene-complexity groups.}
    \label{fig:distribution}
    \vspace{-12pt}
\end{figure}

\noindent\textbf{Reference-Attribute Grouping.} 
To analyze performance across scene conditions, we group scenes using reference textured 3D attributes. For scene $i$, $L_i$ denotes horizontal side length, $t_i$ triangle density, $b_i$ building-footprint coverage, and $r_i^{\mathrm{relief}}$ the difference between the 95th and 5th reference DSM elevation percentiles. 
For scene $i$, $\pi_i(L)$, $\pi_i(t)$, $\pi_i(b)$, and $\pi_i(r^{\mathrm{relief}})$ denote the percentile scores of crop side length, triangle density, building-footprint coverage, and height relief across all scenes, respectively.
We define a composite scene complexity score $\kappa_i$ from these reference attributes by equally weighting scene scale, geometric attributes, and height relief: 
\begin{equation}
\kappa_i=\frac{\pi_i(L)+\tfrac12[\pi_i(t)+\pi_i(b)]+\pi_i(r^{\mathrm{relief}})}{3}.
\label{eq:complexity}
\end{equation}
Sorting scenes by increasing $\kappa_i$ yields Low (184), Medium (183), and High (183) scene complexity groups.
Figure~\ref{fig:distribution} shows their attribute distributions. Attribute computation and grouping details are provided in Section~\ref{sec:supp} of the supplementary material.

\subsection{Appearance and Geometry Evaluation}
\label{sec:metrics}

We evaluate generated appearance and 3D structure using 11 metrics in the two groups of Table~\ref{tab:Sat3D-Bench_4d_height2_full}. Appearance Quality (AQ) covers visual quality and reference appearance agreement. Geometry Fidelity (GF) covers surface agreement, input-shape correspondence, and relative elevation. Detailed metric computation procedures are provided in Section~\ref{sec:supp_metrics} of the supplementary material.

\subsubsection{Appearance Quality}

Appearance evaluation examines both the visual quality of generated scenes and their fidelity to reference appearance. Four metrics use 13 views rendered under shared camera, lighting, background, and color settings. Q-Align-I evaluates complete generated views, while the other three metrics compare corresponding views over reference foreground regions.

\noindent\uline{\textit{Visual Quality (Q-Align-I):}} To assess the overall visual quality of rendered scenes, we use the reference-free image-quality task of Q-Align~\cite{wu2024qalign}. This metric evaluates the generated appearance itself, independently of its agreement with a particular reference scene.

\noindent\uline{\textit{Color Agreement (PSNR):}} To assess the preservation of reference colors and brightness, we use PSNR to compare corresponding rendered views. Following foreground evaluation practice in CO3D~\cite{reizenstein2021co3d}, the comparison focuses on reference scene content, with higher scores indicating smaller pixel-level color errors.

\noindent\uline{\textit{Structure Fidelity (SSIM):}} To assess the preservation of projected building outlines and local texture patterns, we use SSIM~\cite{wang2004ssim}. It measures local image-structure similarity, complementing the pixel-level color agreement assessed by PSNR.

\noindent\uline{\textit{Perceptual Similarity (LPIPS):}} To assess appearance differences beyond pixel-level color and local structure, we use LPIPS~\cite{lpips} to compare learned features of generated and reference views. Lower scores indicate closer perceptual appearance, providing a complementary measure of reference fidelity.

\subsubsection{Geometry Fidelity}

Geometry evaluation examines surface accuracy and coverage, correspondence between generated shapes and input images, and recovery of relative vertical structure. Seven metrics assess these complementary properties using complete normalized surfaces, input--shape pairs, and relative-elevation maps.

\noindent\uline{\textit{Surface Distance (NCD):}} To assess the average spatial deviation between generated and reference surfaces, we use NCD based on the distance computation in the Occupancy Networks evaluation implementation~\cite{mescheder2019occupancy}. It considers distances in both directions, with lower scores indicating closer surface agreement on average.

\noindent\uline{\textit{Surface Matching (F-score):}} To jointly assess generated-surface accuracy and reference-surface coverage, we use F-score following Tanks and Temples~\cite{knapitsch2017tanks}. By combining precision and recall, it measures overall surface matching within a fixed distance tolerance, complementing mean surface distance.

\noindent\uline{\textit{Surface Coverage (Recall):}} To explicitly assess how much of the reference surface is recovered, we separately report Recall following Tanks and Temples~\cite{knapitsch2017tanks}. This metric emphasizes reference-surface coverage, helping distinguish outputs with similar mean distances but different recovered surface extents.

\noindent\uline{\textit{Input-Shape Alignment (ULIP-I):}} To assess global semantic correspondence between generated shapes and input satellite images, we use ULIP-2~\cite{xue2024ulip} through HY3D-Bench~\cite{hy3dBenchMetrics}. The metric compares learned image and shape representations to indicate whether the generated shape agrees with the overall content of the conditioning image.

\noindent\uline{\textit{Input-Shape Alignment (Uni3D-I):}} To assess the same correspondence with an alternative learned representation, we use Uni3D~\cite{zhou2023uni3d} through HY3D-Bench~\cite{hy3dBenchMetrics}. This provides another measure of global image--shape agreement, with the 3D branch representing generated shape rather than texture appearance.

\noindent\uline{\textit{Relative Elevation Error (AbsRel):}} To assess deviations in relative scene heights, we adapt AbsRel from depth evaluation~\cite{eigen2014depth,godard2019monodepth2} to relative-elevation maps. It measures average error relative to reference heights, with lower scores indicating smaller relative-height deviations and missing predicted surfaces retained in the evaluation.

\noindent\uline{\textit{Elevation Accuracy ($\delta_1$):}} To assess how many locations have accurately recovered relative heights, we adopt threshold accuracy $\delta_1$ from depth evaluation~\cite{eigen2014depth,godard2019monodepth2}. It measures the fraction of locations satisfying an elevation-ratio tolerance, distinguishing accurate recovery across more locations from a reduction in average error alone.

Image metrics are first averaged over the 13 views of each scene. All metrics are then averaged across scenes with equal weights and reported separately to distinguish performance across scene properties.

\begin{table*}[!t]
  \centering
  \caption{\textbf{Sat3D-Bench evaluation of 22 generation methods.} Appearance evaluation covers visual quality and reference appearance agreement. Geometry evaluation covers surface agreement, input--shape alignment, and relative elevation. Best and second-best results in each column are marked in \textbf{bold} and \underline{underlined}, respectively.}
  \label{tab:Sat3D-Bench_4d_height2_full}
  \begingroup
  \setlength{\tabcolsep}{1.7pt}
  \renewcommand{\arraystretch}{1.0}
  \resizebox{\textwidth}{!}{%
  \begin{tabular}{@{}l c *{11}{c}@{}}
    \toprule
    \multirow[c]{3}{*}{\textbf{Models}} &
    \multirow[c]{3}{*}{\textbf{Overall $\uparrow$}} &
    \multicolumn{4}{c}{\textbf{Appearance Quality (AQ)}} &
    \multicolumn{7}{c}{\textbf{Geometry Fidelity (GF)}} \\
    \cmidrule(lr){3-6}
    \cmidrule(lr){7-13}
    & & Visual & Pixel & Structure & Perceptual
    & Distance & Matching & Coverage
    & 3D--image & 3D--image
    & Relative Error & Accuracy \\
    & & Q-Align-I$\uparrow$
    & PSNR$\uparrow$
    & SSIM$\uparrow$
    & LPIPS$\downarrow$
    & NCD$\downarrow$
    & F-score$\uparrow$
    & Recall$\uparrow$
    & ULIP-I$\uparrow$
    & Uni3D-I$\uparrow$
    & AbsRel$\downarrow$
    & $\delta_1\uparrow$ \\
    \midrule

    \multicolumn{13}{@{}l}{\emph{\textbf{A. Satellite-to-3D Generation}}} \\

    Sat2Density~[40]
    & 35.95
    & \textbf{2.860} & 7.03 & 0.173 & 0.705
    & \underline{0.042} & \underline{0.186} & \textbf{0.333}
    & 0.079 & 0.114 & 2.215 & \underline{0.152} \\

    Sat2Density++~[41]
    & 34.11
    & 2.192 & 9.86 & 0.206 & 0.515
    & 0.060 & 0.071 & 0.069
    & 0.054 & 0.074 & 3.207 & 0.116 \\

    Sat3DGen~[43]
    & \underline{39.93}
    & 2.337 & \textbf{12.28} & \underline{0.209} & \underline{0.460}
    & \underline{0.042} & 0.100 & 0.124
    & 0.079 & 0.074 & 2.414 & 0.143 \\

    \midrule
    \multicolumn{13}{@{}l}{\emph{\textbf{B. General Image-to-3D Generation}}} \\

    TRELLIS~[62]
    & 38.70
    & 2.231 & 8.02 & 0.153 & 0.558
    & 0.046 & 0.154 & 0.141
    & 0.123 & \textbf{0.191} & \underline{1.042} & 0.107 \\

    TRELLIS.2~[63]
    & 30.46
    & 1.454 & 6.30 & 0.133 & 0.675
    & 0.049 & 0.113 & 0.088
    & 0.080 & 0.183 & 1.074 & 0.051 \\

    Pixal3D~[27]
    & 22.51
    & 1.962 & 5.95 & 0.167 & 0.714
    & 0.204 & 0.030 & 0.026
    & 0.083 & 0.129 & 3.731 & 0.028 \\

    Hunyuan3D-2.1~[22]
    & 24.65
    & 2.220 & \underline{10.41} & 0.164 & 0.582
    & 0.210 & 0.026 & 0.031
    & -0.010 & 0.111 & 24.917 & 0.012 \\

    Step1X-3D~[28]
    & 32.98
    & 2.044 & 7.89 & 0.128 & 0.558
    & 0.071 & 0.135 & 0.150
    & 0.092 & 0.114 & 3.387 & 0.097 \\

    Direct3D-S2~[59]
    & 23.47
    & 2.172 & 6.43 & 0.182 & 0.747
    & 0.158 & 0.067 & 0.098
    & 0.071 & 0.112 & 9.663 & 0.040 \\

    TripoSG~[30]
    & 22.59
    & 2.090 & 6.47 & 0.172 & 0.735
    & 0.175 & 0.049 & 0.096
    & 0.091 & 0.113 & 10.997 & 0.020 \\

    Hi3DGen~[68]
    & 27.97
    & 2.490 & 7.03 & 0.184 & 0.717
    & 0.114 & 0.090 & \underline{0.182}
    & 0.090 & 0.109 & 9.685 & 0.057 \\

    Direct3D~[58]
    & 30.09
    & 2.153 & 6.09 & 0.184 & 0.723
    & 0.070 & 0.112 & 0.119
    & 0.099 & 0.135 & 2.820 & 0.098 \\

    InstantMesh~[66]
    & 19.65
    & 2.399 & 7.00 & 0.170 & 0.696
    & 0.221 & 0.022 & 0.025
    & 0.022 & 0.097 & 12.098 & 0.013 \\

    CRM~[56]
    & 29.90
    & 1.953 & 9.45 & 0.189 & 0.558
    & 0.125 & 0.075 & 0.084
    & 0.052 & 0.085 & 6.335 & 0.073 \\

    SAM 3D Objects~[4]
    & 35.67
    & 2.166 & 9.51 & 0.184 & 0.578
    & 0.048 & 0.126 & 0.131
    & 0.046 & 0.080 & 1.526 & 0.130 \\

    TripoSR~[51]
    & 20.69
    & 2.336 & 6.88 & 0.173 & 0.675
    & 0.157 & 0.033 & 0.034
    & -0.016 & 0.075 & 8.672 & 0.021 \\

    Michelangelo~[74]
    & 24.21
    & \underline{2.835} & 6.21 & 0.194 & 0.796
    & 0.151 & 0.055 & 0.086
    & 0.092 & 0.089 & 7.702 & 0.024 \\

    \midrule
    \multicolumn{13}{@{}l}{\emph{\textbf{C. Scene and World Generation}}} \\

    SceneGen~[36]
    & 36.22
    & 2.212 & 7.85 & 0.142 & 0.571
    & 0.055 & 0.138 & 0.147
    & \textbf{0.129} & 0.163 & 2.100 & 0.094 \\

    WonderWorld~[72]
    & 20.70
    & 1.785 & 6.57 & 0.159 & 0.656
    & 0.223 & 0.032 & 0.026
    & 0.051 & 0.081 & 4.773 & 0.042 \\

    EvoScene~[75]
    & 35.53
    & 2.181 & 7.35 & 0.157 & 0.594
    & 0.054 & 0.153 & 0.134
    & 0.065 & 0.181 & 1.056 & 0.116 \\

    SynCity~[13]
    & 38.48
    & 2.220 & 8.07 & 0.147 & 0.555
    & 0.045 & 0.152 & 0.147
    & \underline{0.128} & \underline{0.184} & 1.305 & 0.104 \\

    Extend3D~[71]
    & \textbf{44.44}
    & 2.219 & 10.10 & \textbf{0.219} & \textbf{0.415}
    & \textbf{0.036} & \textbf{0.193} & 0.179
    & 0.099 & 0.157 & \textbf{0.758} & \textbf{0.154} \\

    \bottomrule
  \end{tabular}%
  }
  \endgroup
\end{table*}

\section{Experiments}
\label{sec:experiments}
This section evaluates 22 generation methods on Sat3D-Bench in terms of appearance quality and geometry fidelity. We compare their performance across scene complexity and scale, and examine how visual quality relates to surface recovery, input--shape alignment, and relative elevation. These comparisons identify limitations of existing methods and directions for improving urban scene generation.

\subsection{Experimental Setup}
\label{sec:exp_setup}

\noindent\textbf{Evaluated Methods.}
We evaluate 22 generation methods on Sat3D-Bench, grouped into three categories, i.e., satellite-to-3D generation, general image-to-3D generation, and scene and world generation, as summarized in Table~\ref{tab:Sat3D-Bench_4d_height2_full}. The three satellite-to-3D methods are Sat2Density~\cite{qian2023sat2density}, Sat2Density++~\cite{sat2densitypp}, and Sat3DGen~\cite{sat3dgen}. The 14 general image-to-3D methods comprise TRELLIS~\cite{xiang2025trellis}, TRELLIS.2~\cite{xiang2026trellis2}, Pixal3D~\cite{li2026pixal3d}, Hunyuan3D-2.1~\cite{hunyuan3d2025v21},
Step1X-3D~\cite{li2025step1x3d}, Direct3D-S2~\cite{wu2025direct3ds2}, TripoSG~\cite{li2025triposg}, Hi3DGen~\cite{ye2025hi3dgen}, Direct3D~\cite{wu2024direct3d},
InstantMesh~\cite{xu2024instantmesh}, CRM~\cite{wang2024crm}, SAM 3D Objects~\cite{sam3dteam2025sam3d3dfyimages}, TripoSR~\cite{tochilkin2024triposr}, and Michelangelo~\cite{zhao2023michelangelo}. The five scene and world generation methods are
SceneGen~\cite{meng2025scenegen}, WonderWorld~\cite{yu2025wonderworld}, EvoScene~\cite{zheng2025self}, SynCity~\cite{engstler2025syncity}, and Extend3D~\cite{yoon2026extend3d}. SynCity uses its single-image reconstruction workflow, and WonderWorld is evaluated through its derived mesh.

\begin{figure*}[t]
    \centering
    \includegraphics[width=.97\linewidth]{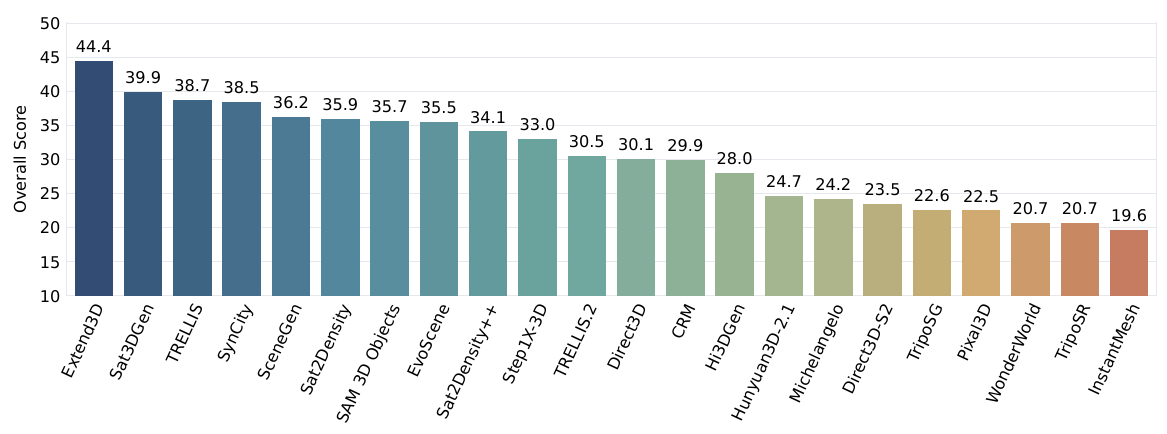}
    \vspace{-6pt}
    \caption{\textbf{Overall ranking on Sat3D-Bench.} The 22 methods are ranked by their overall score across 11 appearance and geometry metrics. Extend3D ranks first with 44.4 out of 100.}
    \label{fig:benchmark_comparison}
    \vspace{-10pt}
\end{figure*}

\begin{figure}[t]
    \centering
    \includegraphics[width=.95\linewidth]{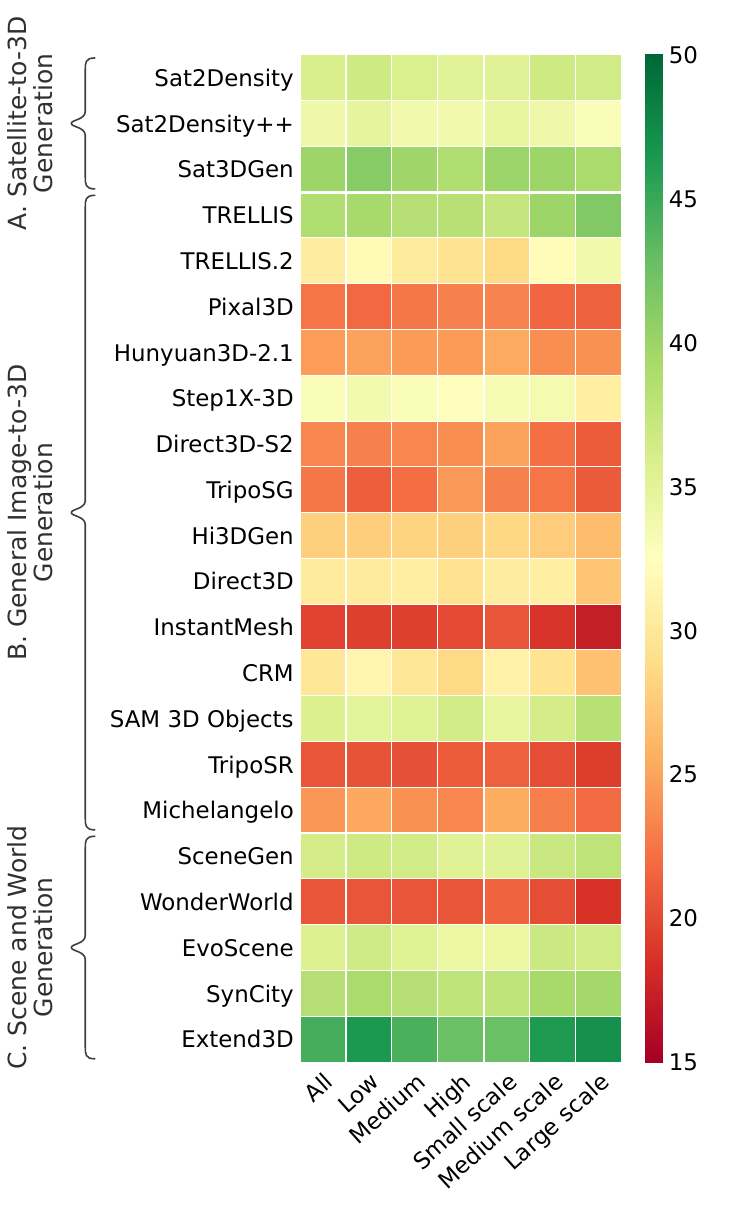}
    \vspace{-10pt}
    \caption{\textbf{Overall scores across scene complexity and scale.} Each cell averages 11 metrics normalized to $[0,100]$ using shared mappings following WorldScore~\cite{duan2025worldscore}. Higher is better.}
    \label{fig:score_heatmap}
    \vspace{-10pt}
\end{figure}

\noindent\textbf{Data and Evaluation.}
The evaluation set comprises 550 pairs of satellite images and textured 3D scenes from 29 cities. All experiments are conducted on 8 NVIDIA RTX 4090 GPUs. We evaluate all methods using the 11 metrics described in Sections~\ref{sec:task} and~\ref{sec:metrics}. For image-based metrics, scores are first averaged over the 13 views of each scene. Each metric is then averaged across evaluated scenes with equal scene weights. Reported means cover all 550 scenes for most methods. For Pixal3D and EvoScene, they cover the successfully evaluated subsets of 540 and 534 scenes, respectively. 
To facilitate comparison across appearance quality and geometry fidelity, Figure~\ref{fig:benchmark_comparison} reports an overall score for each method.
We normalize the 11 metrics to $[0,100]$ and then average the resulting scene scores. Metric normalization and overall score calculation are detailed in Section~\ref{sec:supp_score_normalization} of the supplementary material.

\subsection{Analysis}

Table~\ref{tab:Sat3D-Bench_4d_height2_full} compares the 22 methods using individual appearance and geometry metrics. 
Figure~\ref{fig:benchmark_comparison} ranks the 22 methods by overall score, with Extend3D ranking first at 44.4 out of 100. Figure~\ref{fig:score_heatmap} summarizes their overall scores across scene complexity and scale. We analyze these results to characterize the capabilities and limitations of existing methods and discuss directions for improving urban scene generation.

\noindent\textbf{Performance across Scene Complexity and Scale.}
In Table~\ref{tab:Sat3D-Bench_4d_height2_full}, Sat3DGen~\cite{sat3dgen} achieves the highest PSNR, and Sat2Density~\cite{qian2023sat2density} achieves the highest surface recall. Extend3D~\cite{yoon2026extend3d} obtains the highest overall score in every complexity and scale group in Figure~\ref{fig:score_heatmap}, whereas InstantMesh~\cite{xu2024instantmesh} and
TripoSR~\cite{tochilkin2024triposr} obtain low surface matching and coverage scores in Table~\ref{tab:Sat3D-Bench_4d_height2_full}. Differences in training data and scene representation may contribute to these performance patterns. For example, Sat2Density learns scene geometry from satellite--ground image pairs, whereas general asset generators primarily target object-level inputs and outputs. Extend3D adapts object-level priors through spatial latent expansion. Figure~\ref{fig:score_heatmap} further shows generally lower overall scores
for high-complexity scenes, while performance trends across scene scales differ even within the same method category. TRELLIS~\cite{xiang2025trellis} and Extend3D obtain higher
scores for large-scale scenes than for small-scale scenes, whereas InstantMesh and WonderWorld~\cite{yu2025wonderworld} show the opposite trend. These observations motivate further investigation of adaptation to satellite imagery and representations that jointly organize multiple scene structures.

\noindent\textbf{Appearance and Geometry Fidelity.}  Better visual quality, reference appearance, or global input--shape alignment does not necessarily coincide with better geometric fidelity. Among satellite-to-3D methods, Sat3DGen~\cite{sat3dgen} achieves better reference appearance
than Sat2Density~\cite{qian2023sat2density}, whereas Sat2Density attains higher Q-Align-I, F-score, and Recall. Among general image-to-3D and scene and world generation methods, TRELLIS~\cite{xiang2025trellis} and SceneGen~\cite{meng2025scenegen} achieve the highest Uni3D-I and ULIP-I scores, respectively, while Extend3D~\cite{yoon2026extend3d} outperforms both in reference appearance, surface matching, and relative elevation. These comparisons show that rendered appearance and global input--shape alignment alone are insufficient to characterize faithful recovery of the corresponding urban environment. Future work could explore local image--structure correspondence to improve reference appearance while preserving geometric fidelity.

\noindent\textbf{Structural Accuracy and Completeness.} The geometric results in Table~\ref{tab:Sat3D-Bench_4d_height2_full}
show that average error alone does not fully characterize accurate scene recovery. Sat2Density~\cite{qian2023sat2density} and
Sat3DGen~\cite{sat3dgen} share a reported NCD of $0.042$, but their Recall values are $0.333$ and $0.124$,
respectively, indicating different reference-surface coverage despite similar mean distances. Likewise, TRELLIS~\cite{xiang2025trellis} achieves lower
AbsRel than Sat3DGen but also lower $\delta_1$, indicating smaller average relative elevation error but a lower proportion of accurately recovered heights. These comparisons support jointly reporting average errors, surface coverage, and elevation accuracy.

These findings motivate combining adaptation to satellite observations, scene-level representations, local image-structure correspondence, and objectives for geometric accuracy and completeness. Future methods should be evaluated using complementary appearance and geometry metrics across scene complexity and scale.

\section{Conclusion}
\label{sec:conclusion}

This paper presented Sat3D-Bench for evaluating large-scale 3D scene generation from a single satellite image. The benchmark provides 550 pairs of high-quality satellite images and textured 3D scenes from 29 cities across five continents, together with a unified evaluation framework
for appearance quality and geometry fidelity. Comparisons of 22 generation methods show that higher visual quality does not necessarily imply faithful scene reconstruction, while similar mean surface distances can accompany substantial differences in surface coverage. These findings motivate future research on joint modeling of global layout and local geometry, spatially explicit image conditioning, and recovery of relative vertical structure. Sat3D-Bench provides a common evaluation basis for measuring progress toward these goals.

{
\small
\bibliographystyle{ieeenat_fullname}
\bibliography{main}
}

\ifdefined\SatThreeDMainDocument
\else
\documentclass[10pt,twocolumn,letterpaper]{article}
\usepackage[review]{cvpr}
\usepackage{times,graphicx,amsmath,amssymb,booktabs,tabularx,array}
\usepackage[normalem]{ulem}
\usepackage{stfloats}
\usepackage{multirow}

\newcommand{\red}[1]{{\color{red}#1}}
\newcommand{\todo}[1]{{\color{red}#1}}
\newcommand{\TODO}[1]{\textbf{\color{red}[TODO: #1]}}






\definecolor{cvprblue}{rgb}{0.21,0.49,0.74}
\usepackage[pagebackref,breaklinks,colorlinks,allcolors=cvprblue]{hyperref}
\def\confName{CVPR}
\def\confYear{2027}
\title{Sat3D-Bench: Benchmarking Large-Scale 3D Scene Generation \\ from a Single Satellite Image}
\author{Anonymous CVPR submission}
\newcommand{\SatThreeDStandaloneBegin}{\begin{document}}
\newcommand{\SatThreeDStandaloneEnd}{\end{document}}
\fi
\ifdefined\SatThreeDStandaloneBegin
  \expandafter\SatThreeDStandaloneBegin
\fi

\clearpage
\ifdefined\microtypesetup
  \microtypesetup{activate=false}
\fi
\ifPDFTeX
  \renewcommand{\encodingdefault}{OT1}
  \fontencoding{OT1}\selectfont
\fi
\appendix
\setcounter{section}{0}
\setcounter{figure}{0}
\setcounter{table}{0}
\setcounter{equation}{0}

\renewcommand{\thefigure}{S\arabic{figure}}
\renewcommand{\thetable}{S\arabic{table}}
\renewcommand{\theequation}{S\arabic{equation}}

\renewcommand{\theHsection}{supp.\Alph{section}}
\renewcommand{\theHfigure}{supp.\arabic{figure}}
\renewcommand{\theHtable}{supp.\arabic{table}}
\renewcommand{\theHequation}{supp.\arabic{equation}}

\maketitlesupplementary
\setcounter{dbltopnumber}{1}

\begin{table*}[!b]
\centering
\caption{\textbf{Generation settings and exported outputs for the 22 evaluated methods.}
ArXiv Date gives the first arXiv submission date.
Resolution gives the mesh extraction grid or native representation size, reported per patch for EvoScene and Extend3D.
Parameter counts include frozen encoders and texture networks.
Time is the mean inference time.
For Pixal3D ($\dagger$), 538 outputs used a $1536^3$ grid and two used a $1408^3$ grid.
``UV'' denotes texture maps, ``VC'' denotes vertex colors, ``PBR'' denotes physically based materials, and ``Geom.'' denotes untextured geometry.
Adaptive indicates a variable representation size.}
\small
\setlength{\tabcolsep}{2.5pt}
\renewcommand{\arraystretch}{1.}
\begin{tabular*}{0.93\textwidth}{@{\extracolsep{\fill}}lccrrl@{}}
\toprule
Method & ArXiv Date
 & Resolution & Params (M) & Time (s) & Output \\
\midrule
\multicolumn{6}{@{}l}{\emph{\textbf{A. Satellite-to-3D Generation}}} \\
Sat2Density~\cite{qian2023sat2density} & 23.03.26 & $256^2\!\times\!65$ & 31.6 & 1.38 & GLB (Geom.) \\
Sat2Density++~\cite{sat2densitypp} & 25.05.22 & $256^3$ & 27.0 & 8.44 & OBJ (VC) \\
Sat3DGen~\cite{sat3dgen} & 26.05.14 & $320^3$ & 373.9 & 13.22 & OBJ (VC) \\
\midrule
\multicolumn{6}{@{}l}{\emph{\textbf{B. General Image-to-3D Generation}}} \\
TRELLIS~\cite{xiang2025trellis} & 24.12.02 & $256^3$ & 1,710.3 & 30.71 & GLB (UV) \\
TRELLIS.2~\cite{xiang2026trellis2} & 25.12.16 & $512^3$ & 5,202.0 & 157.94 & GLB (PBR) \\
Pixal3D~\cite{li2026pixal3d}$^{\dagger}$ & 26.05.11 & $1536^3\!/1408^3$ & 7,736.1 & 326.62 & GLB (PBR) \\
Hunyuan3D-2.1~\cite{hunyuan3d2025v21} & 25.06.18 & $384^3$ & 6,768.4 & 88.64 & GLB (PBR) \\
Step1X-3D~\cite{li2025step1x3d} & 25.05.12 & $384^3$ & 6,110.7 & 196.30 & GLB (UV) \\
Direct3D-S2~\cite{wu2025direct3ds2} & 25.05.23 & $1024^3$ & 2,616.5 & 183.12 & GLB (Geom.) \\
TripoSG~\cite{li2025triposg} & 25.02.10 & $512^3$ & 1,958.1 & 27.07 & GLB (Geom.) \\
Hi3DGen~\cite{ye2025hi3dgen} & 25.03.28 & $256^3$ & 2,937.7 & 8.79 & GLB (Geom.) \\
Direct3D~\cite{wu2024direct3d} & 24.05.23 & $512^3$ & 1,610.9 & 12.88 & GLB (Geom.) \\
InstantMesh~\cite{xu2024instantmesh} & 24.04.10 & $128^3$ & 2,300.7 & 151.72 & OBJ (UV) \\
CRM~\cite{wang2024crm} & 24.03.08 & $80^3$ & 4,111.8 & 32.38 & GLB (UV) \\
SAM 3D Objects~\cite{sam3dteam2025sam3d3dfyimages} & 25.11.20 & $256^3$ & 3,385.1 & 37.20 & GLB (UV) \\
TripoSR~\cite{tochilkin2024triposr} & 24.03.04 & $256^3$ & 419.3 & 1.75 & GLB (VC) \\
Michelangelo~\cite{zhao2023michelangelo} & 23.06.29 & $128^3$ & 528.5 & 19.17 & OBJ (Geom.) \\
\midrule
\multicolumn{6}{@{}l}{\emph{\textbf{C. Scene and World Generation}}} \\
SceneGen~\cite{meng2025scenegen} & 25.08.21 & $256^3$ & 2,972.7 & 38.81 & GLB (UV) \\
WonderWorld~\cite{yu2025wonderworld} & 24.06.13 & Adaptive & 2,579.9 & 3.69 & PLY (3DGS) \\
EvoScene~\cite{zheng2025self} & 25.12.09 & $256^3$/patch & 16,522.3 & 3,414.90 & GLB (UV) \\
SynCity~\cite{engstler2025syncity} & 25.03.20 & $256^3$ & 1,710.3 & 33.58 & GLB (UV) \\
Extend3D~\cite{yoon2026extend3d} & 26.03.31 & $256^3$/patch & 2,101.2 & 724.08 & GLB (UV) \\
\bottomrule
\end{tabular*}
\label{tab:generation-method-details}
\end{table*}

\section{Unified Representation Adapter Details}
\label{sec:supp_adapters}

\subsection{Mesh Conversion and Orientation}
\label{sec:supp_alignment}

Following the evaluation procedure in Section~\ref{sec:task} of the main paper, we convert generated outputs into meshes when needed. This allows us to render all generated scenes with the same camera and evaluate them using the same surface metrics. Table~\ref{tab:generation-method-details} lists the generation settings and output formats for all 22 methods. For Sat2Density~\cite{qian2023sat2density}, we use the mesh extracted from its predicted density field. For WonderWorld~\cite{yu2025wonderworld}, we construct a textured mesh from color and depth images rendered by its native Gaussian renderer. For methods that already provide meshes, including EvoScene~\cite{zheng2025self}, we directly use their generated scenes. We import all meshes into Blender~\cite{blender}, preserving all scene components, their positions, rotations, and scales, and any available textures, materials, or vertex colors.

We then orient the meshes according to the coordinate convention used for evaluation, with the positive z-axis pointing upward. All scenes produced by the same method follow one fixed orientation rule. The paired textured references are reserved for evaluation and are neither provided to the generators nor used to adjust the orientation of individual predictions.

\subsection{Independent Spatial Normalization}

After standardizing orientation, we normalize the generated mesh and its reference separately, following Eq.~\eqref{eq:independent_scene_norm} of the main paper. For each mesh, we first center its horizontal bounding box at the horizontal coordinate origin and shift it vertically until its lowest point has height zero. We then uniformly scale the entire mesh so that the larger of its two horizontal spans equals one. Applying the same scale factor in all three directions preserves the original proportions.

The prediction and reference each use their own bounds to determine the translation and scale. This places them at a common horizontal scale for comparing building layouts, building sizes relative to the scene, and relative heights. The resulting meshes are used for subsequent multi-view rendering, surface comparison, and relative-elevation evaluation.

\begin{figure*}[!b]
\centering\includegraphics[width=\textwidth]{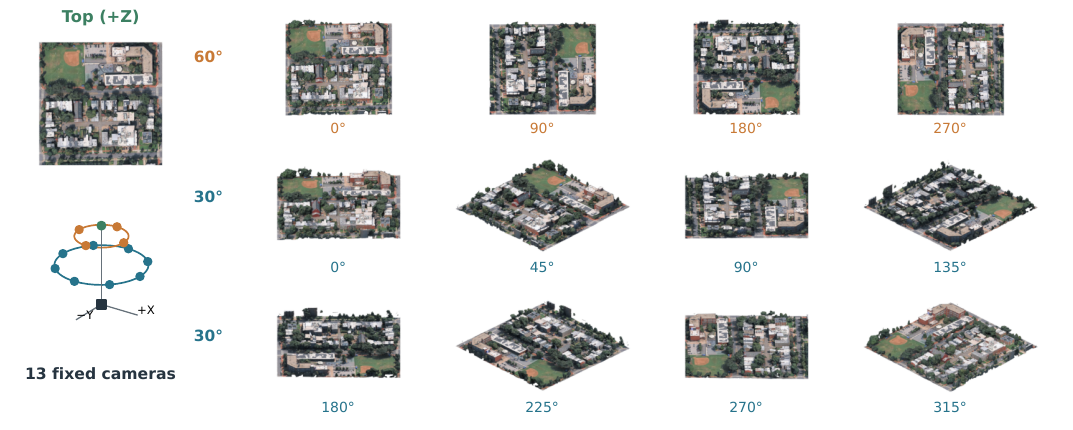}
\caption{\textbf{Camera directions for appearance evaluation.} The Manhattan reference is shown from above and from two rings of oblique views. Matching colors identify the camera for each rendering. The labels give elevation and azimuth. The display crops are shared within each ring.}
\label{fig:supp_views}
\end{figure*}

\section{Reference-Attribute Grouping Details}
\label{sec:supp}

Following Section~\ref{sec:curation} of the main paper, we use four attributes of the reference scenes to define complexity groups, including scene size, triangle density, building coverage, and height relief. Scene size is the side length of the square crop in metres. Triangle density is the number of triangles in the exported reference mesh divided by the crop area. Building coverage is the ground area occupied by buildings within the crop divided by the crop area. Height relief is the difference between the 95th and 5th percentiles of valid elevation values in the reference digital surface model (DSM). We calculate this difference in metres before mesh normalization.

To place these attributes on a common scale, we convert
their raw values to rank-based percentile scores. For each attribute, we rank all 550 scenes in ascending order, assigning average ranks to tied values. The percentile score is obtained by subtracting 0.5 from the rank and dividing by 550.

We first average the percentile scores for triangle density and building coverage to obtain a combined score for geometric attributes. We then equally average this score with the percentile
scores for scene size and height relief to obtain the complexity score in Eq.~\eqref{eq:complexity} of the main paper.

We sort scenes from the lowest to the highest complexity score. For equal scores, we sort scenes by scene ID. We then divide the ordered scenes into Low, Medium, and High groups containing 184, 183, and 183 scenes, respectively. Figure~\ref{fig:distribution} of the main paper shows the distributions of the scores for scene size, geometric attributes, and height relief across these groups in panels (a), (b), and (c), respectively.

To compare performance across different scene sizes, we also group scenes by their physical size. The Small group covers side lengths from 100 to 200 m, the Medium group covers lengths greater than 200 m and up to 400 m, and the Large group covers lengths greater than 400 m and up to 1,000 m. Figure~\ref{fig:score_heatmap} of the main paper reports results for both the complexity and scale groups.

\section{Evaluation Metrics Details}
\label{sec:supp_metrics}

This section describes the calculation of the metrics introduced in Section~\ref{sec:metrics} of the main paper. The code for computing these metrics will be publicly released upon acceptance.

\subsection{Appearance Quality}
\label{sec:supp_aq}

We render each normalized generated 3D scene and its reference from the same 13 fixed views, using identical camera, lighting, color, and background settings (Figure~\ref{fig:supp_views}). Q-Align-I~\cite{wu2024qalign} evaluates each complete generated view by averaging quality levels according to their predicted probabilities. 

Following CO3D~\cite{reizenstein2021co3d}, we use a foreground mask rendered from the reference 3D model to exclude irrelevant background regions when computing PSNR~\cite{reizenstein2021co3d}, SSIM~\cite{wang2004ssim}, and LPIPS~\cite{lpips}. This focuses the comparison between generated and reference views on the scene itself. SSIM~\cite{wang2004ssim} compares brightness, contrast, and structure in corresponding image windows, then averages the similarities over windows touching the reference foreground. LPIPS~\cite{lpips} compares AlexNet features, averages feature differences within each layer using reference-foreground coverage as weights, and adds the layer results. For SSIM and LPIPS, we first compute local similarities or feature differences from the complete images, then average these local results using the same reference mask.

\subsection{Geometry Fidelity}
\label{sec:supp_gf}

We sample points from the complete normalized prediction and reference meshes in proportion to triangle area. 
To compute NCD, we find the nearest reference point for each predicted point and average their Euclidean distances. We repeat this calculation from reference to prediction, then average the two directional results~\cite{mescheder2019occupancy}. F-score and Recall determine point matches using the same fixed distance threshold. Precision measures the fraction of predicted points close to the reference surface, while Recall measures the fraction of reference points covered by the prediction. F-score combines these fractions using their harmonic mean~\cite{knapitsch2017tanks}.

To compare the input image with the generated 3D scene, we sample its surface to obtain a point cloud following HY3D-Bench~\cite{hy3dBenchMetrics}. ULIP-I and Uni3D-I extract the input image and point cloud features using ULIP-2~\cite{xue2024ulip} and Uni3D~\cite{zhou2023uni3d}, respectively. For each model, we use cosine similarity to compare the features of the input image and the generated point cloud.

For relative-elevation evaluation, we extract elevation maps from the predicted and reference scenes. At the same horizontal locations in both scenes, we record the height of the uppermost surface, such as a roof or road. We use each scene's lowest point as zero. We evaluate locations where the reference surface is present and its relative elevation exceeds the fixed minimum threshold. Missing predictions receive elevation zero and remain included. Following depth-evaluation definitions~\cite{eigen2014depth,godard2019monodepth2}, AbsRel averages absolute elevation errors divided by the corresponding reference elevations. Elevation accuracy ($\delta_1$) is the fraction of evaluated locations
with a positive predicted elevation and a predicted-to-reference
elevation ratio strictly between $1/1.25$ and $1.25$.

\section{Experimental Setup Details}
\label{sec:supp_validation}

SynCity~\cite{engstler2025syncity} follows its single-image reconstruction workflow.  Experiments were conducted on 8 NVIDIA RTX 4090 GPUs. 20 methods were evaluated on all 550 scenes. For Pixal3D~\cite{li2026pixal3d} and EvoScene~\cite{zheng2025self}, the reported averages were calculated from 540 and 534 successfully evaluated scenes, respectively.

To check the metric calculations, we compared identical image pairs and identical point-set pairs. As expected, the image pairs gave an SSIM~\cite{wang2004ssim} of one, and the point-set pairs gave an NCD~\cite{mescheder2019occupancy} of zero. We also checked the foreground-based appearance results against independent calculations and obtained matching results.

\begin{figure}[htbp]
    \centering
    \includegraphics[width=1.\linewidth]{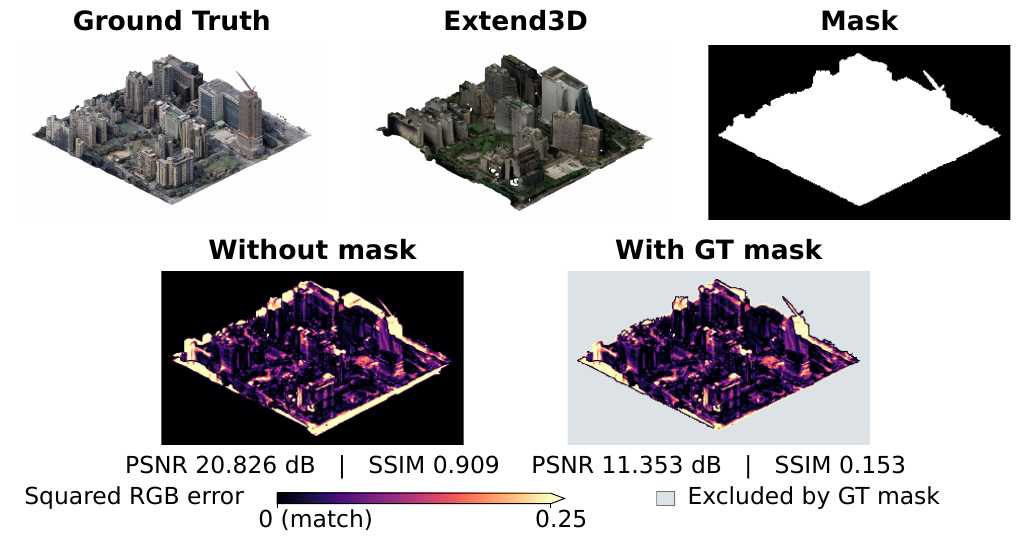}
    \caption{\textbf{Effect of the mask on appearance scores.} Scores computed with and without the reference mask are compared for the same oblique view of a Taipei scene generated by Extend3D~\cite{yoon2026extend3d}. The lower panels show the same squared RGB errors using the same color scale, with areas excluded by the reference mask shown in gray.}
    \label{fig:supp_mask_comparison}
\end{figure}

To illustrate the effect of foreground-based aggregation, we compare full-frame and masked scores for a single oblique view of a Taipei scene generated by Extend3D~\cite{yoon2026extend3d} (Fig.~\ref{fig:supp_mask_comparison}). Applying the reference mask decreases PSNR from 20.826 to 11.353~dB and SSIM from 0.909 to 0.153. The mask reduces the contribution of the shared background, giving differences in the appearance of buildings and streets greater weight in the scores.

\section{Overall Score Calculation Details}
\label{sec:supp_raw_means}
\label{sec:supp_score_normalization}
\label{sec:supp_calibration}
\label{sec:supp_aggregation}

Table~\ref{tab:Sat3D-Bench_4d_height2_full} of the main paper reports the results of all 22 methods on the 11 appearance and geometry metrics. Each metric evaluates a different aspect of the generated scenes, and the overall score combines these metrics to compare overall performance.

To make differences in overall performance easier to see, Figure~\ref{fig:benchmark_comparison} shows the methods in order from highest to lowest overall score. Extend3D~\cite{yoon2026extend3d} ranks first with 44.4 out of 100, followed by Sat3DGen~\cite{sat3dgen} with 39.9. The following paragraphs explain how we calculate the individual metric results and overall scores.

We start by evaluating each generated scene separately for each method. For appearance metrics, we calculate a value for each of the 13 rendered views and average these values to obtain the result for that scene. Geometry metrics are computed directly for each scene. We then average each metric across the evaluated scenes, giving every scene equal weight. Table~\ref{tab:Sat3D-Bench_4d_height2_full} of the main paper reports these averages on each metric's original scale.

To calculate the overall scores, we use the metric values for each scene before averaging across scenes. The metrics have different numerical scales, and better performance corresponds to larger values for some metrics and smaller values for others. We therefore convert them to scores from 0 to 100, with higher scores always indicating better performance. Following the normalization principle of WorldScore~\cite{duan2025worldscore}, we use the same conversion rule for each metric across all methods and groups.

\begin{figure*}[!t]
    \centering
    \includegraphics[width=\textwidth]{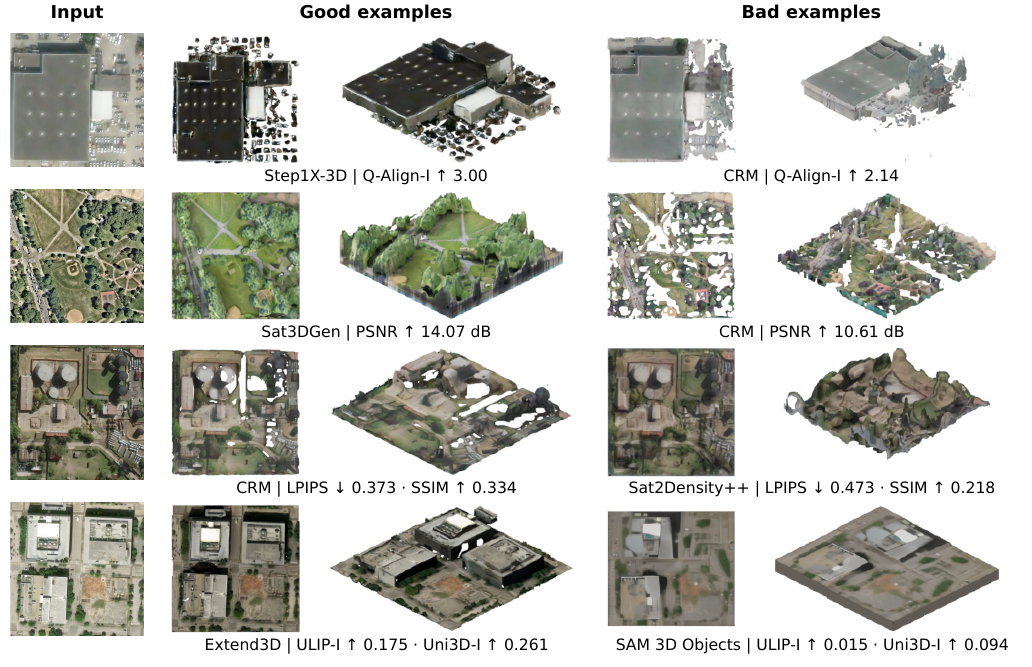}
    \caption{\textbf{Appearance and input--shape comparisons.} Each row shows a satellite input and two generated scenes in top and oblique views. The first row illustrates visual quality, the next two illustrate agreement with the paired textured reference, and the final row illustrates correspondence between the input image and the generated shape. Appearance scores are averaged over 13 evaluation views. Arrows indicate the preferred direction.}
    \label{fig:supp_appearance_examples}
\end{figure*}

\begin{figure*}[!t]
    \centering
    \includegraphics[width=\textwidth]{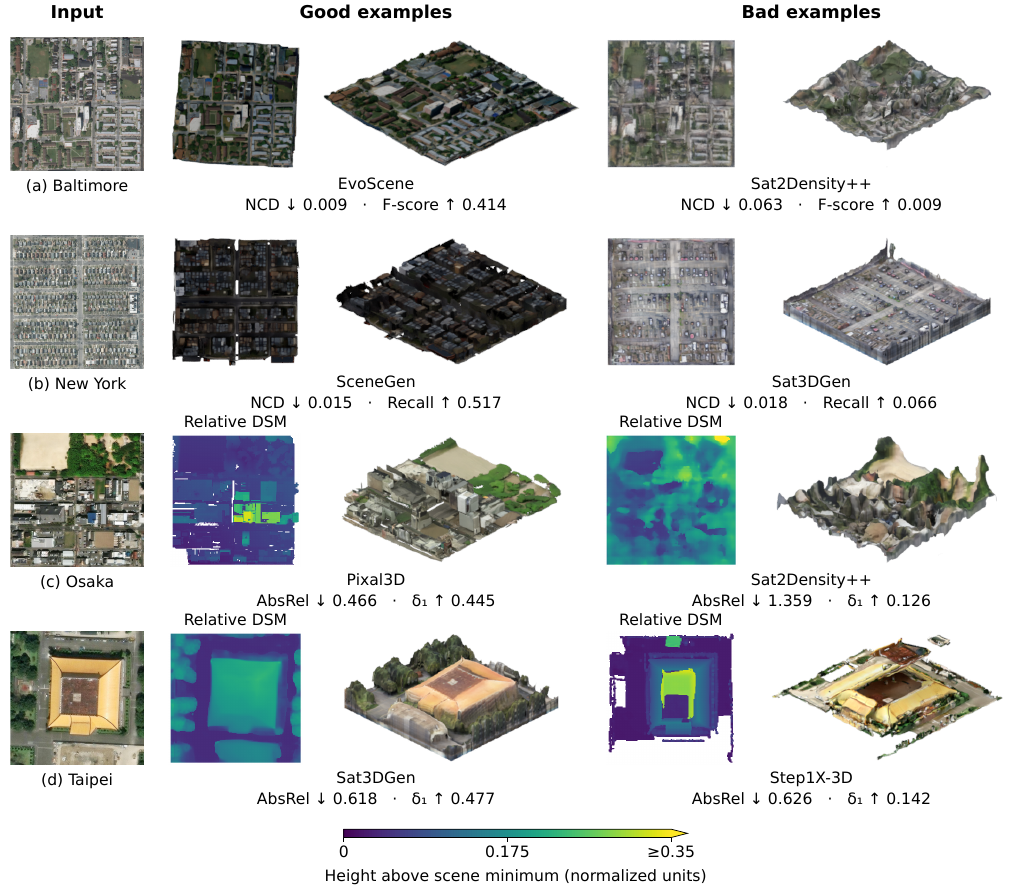}
    \caption{\textbf{Surface geometry and relative-elevation comparisons.} Each row compares two generated scenes for the same satellite input. Panels (a) and (b) illustrate surface distance, matching, and coverage. Panels (c) and (d) show relative DSMs and oblique views. The DSMs use a shared color scale to show elevation above each normalized scene's minimum. Arrows indicate the preferred direction.}
    \label{fig:supp_geometry_examples}
\end{figure*}

For some metrics, the conversion range is determined from the evaluation results. For each of these metrics, we collect the valid scene-level values from all 22 methods and give every method the same total weight. Each method's weight is shared equally among its valid scene values. We use the weighted 5th and 95th percentiles as calibration points for PSNR, ULIP-I, and Uni3D-I. For LPIPS and NCD, we use zero and the weighted 95th percentile as calibration points.

Once the conversion rules are set, we apply them to all valid scene-level metric values, including those outside the calibration range. Any converted score below zero is set to zero, and any score above 100 is set to 100. For SSIM~\cite{wang2004ssim}, negative scene-level values are retained when calculating the original metric averages but receive a score of zero when calculating the overall scores.

With all metrics on the same scale, we retain scenes with valid results for all 11 metrics. For each scene, we average the 11 metric scores with equal weights to obtain a scene score. We then average these scene scores, again giving every scene equal weight, to obtain each method's overall score shown in Figure~\ref{fig:benchmark_comparison} of the main paper.

To compare performance across scene complexity and scale, we also average the scene scores separately within each group defined in Section~\ref{sec:supp}. All groups use the same metric conversion rules. Figure~\ref{fig:score_heatmap} of the main paper presents these group-level results.
The code implementing the metric-specific mappings and overall score calculation will be publicly released upon acceptance.

\section{Qualitative Evaluation Examples}
\label{sec:supp_qualitative}

Figure~\ref{fig:supp_appearance_examples} illustrates visual quality, reference appearance agreement, and input–shape alignment. Figure~\ref{fig:supp_geometry_examples} illustrates surface geometry and relative elevation. Each row compares scenes generated by two methods from the same satellite image, illustrating differences in the evaluated property. All outputs are processed following the procedure described in Section~\ref{sec:supp_adapters}. Top views illustrate scene layouts, while oblique views reveal building geometry and vertical structure. The reported appearance metric scores are averaged over all 13 evaluation views for each scene.

The top and oblique views in Figure~\ref{fig:supp_appearance_examples} reveal differences in scene completeness, roof and road patterns, and building volume. The accompanying scores connect these visible differences to the properties evaluated by the corresponding metrics. In the final row, Extend3D~\cite{yoon2026extend3d} produces distinct three-dimensional building volumes, while the SAM 3D Objects~\cite{sam3dteam2025sam3d3dfyimages} result is much flatter. This difference is clearer in the oblique views.

The first two rows of Figure~\ref{fig:supp_geometry_examples} illustrate the value of reporting surface distance and coverage together. In the New York example, SceneGen~\cite{meng2025scenegen} and Sat3DGen~\cite{sat3dgen} have similar NCD~\cite{mescheder2019occupancy} values, but their Recall~\cite{knapitsch2017tanks} values are 0.517 and 0.066, respectively. Reporting both metrics distinguishes outputs with similar average surface distances but substantially different reference-surface coverage.

The last two rows of Figure~\ref{fig:supp_geometry_examples} reveal differences in vertical structure through relative DSMs and oblique views. All DSMs use the same color scale, with values at or above 0.35 sharing the upper-end color. In the Taipei example, the two methods have similar average relative-elevation errors~\cite{eigen2014depth,godard2019monodepth2}, but Sat3DGen~\cite{sat3dgen} recovers elevations close to the reference at more evaluated locations. Its elevation accuracy~\cite{eigen2014depth,godard2019monodepth2} is 47.7\%, compared with 14.2\% for Step1X-3D~\cite{li2025step1x3d}.

\ifdefined\SatThreeDStandaloneEnd
  \expandafter\SatThreeDStandaloneEnd
\fi

\end{document}